\documentclass[10pt,twocolumn,letterpaper]{article}

\usepackage{cvpr}              

\usepackage{rotating}
\usepackage{multirow}

\usepackage[linesnumbered,ruled,vlined]{algorithm2e}
\usepackage{algorithmic} %

\definecolor{cvprblue}{rgb}{0.21,0.49,0.74}
\usepackage[pagebackref,breaklinks,colorlinks,allcolors=cvprblue]{hyperref}

\title{Completion as Enhancement: \\A Degradation-Aware Selective Image Guided Network for Depth Completion}

\author{Zhiqiang Yan \quad Zhengxue Wang \quad Kun Wang \quad Jun Li\thanks{Corresponding authors} \quad Jian Yang\footnotemark[1]\\
PCA Lab\thanks{PCA Lab, Key Lab of Intelligent Perception and Systems for High-Dimensional Information of Ministry of Education, and Jiangsu Key Lab of Image and Video Understanding for Social Security, School of Computer Science and Engineering, Nanjing University of Sci. \& Tech.}, \ 
Nanjing University of Science and Technology\\
{\tt\small \{yanzq,zxwang,kunwang,junli,csjyang\}@njust.edu.cn}
}

\begin{document}
\maketitle
\begin{abstract}

In this paper, we introduce the \textbf{S}elective \textbf{I}mage \textbf{G}uided \textbf{Net}work (SigNet), a novel degradation-aware framework that transforms depth completion into depth enhancement for the first time. 
Moving beyond direct completion using convolutional neural networks (CNNs), SigNet initially densifies sparse depth data through non-CNN densification tools to obtain coarse yet dense depth. This approach eliminates the mismatch and ambiguity caused by direct convolution over irregularly sampled sparse data. 
Subsequently, SigNet redefines completion as enhancement, establishing a self-supervised degradation bridge between the coarse depth and the targeted dense depth for effective RGB-D fusion. 
To achieve this, SigNet leverages the implicit degradation to adaptively select high-frequency components (e.g., edges) of RGB data to compensate for the coarse depth. This degradation is further integrated into a multi-modal conditional Mamba, dynamically generating the state parameters to enable efficient global high-frequency information interaction. 
We conduct extensive experiments on the NYUv2, DIML, SUN RGBD, and TOFDC datasets, demonstrating the state-of-the-art (SOTA) performance of SigNet.

\end{abstract}    
\section{Introduction}
\label{sec:intro}
Depth completion \cite{ma2018sparse,yan2022rignet,tang2024bilateral,yan2024tri} aims to predict a dense depth map from sparse and noisy depth measurements, often using the corresponding RGB image for auxiliary guidance. The resulting dense depth map can benefit various downstream applications such as scene reconstruction \cite{thomas2013flexible,yan2022multi,song2015sun,zhu2025voxelsplat} and self-driving \cite{liu2021fcfr,yan2023learnable,zhu2023curricular,wang2024improving,zhou2023unsupervised,zhou2024adverse}. 

\begin{figure}[t]
  \centering
  \includegraphics[width=0.97\columnwidth]{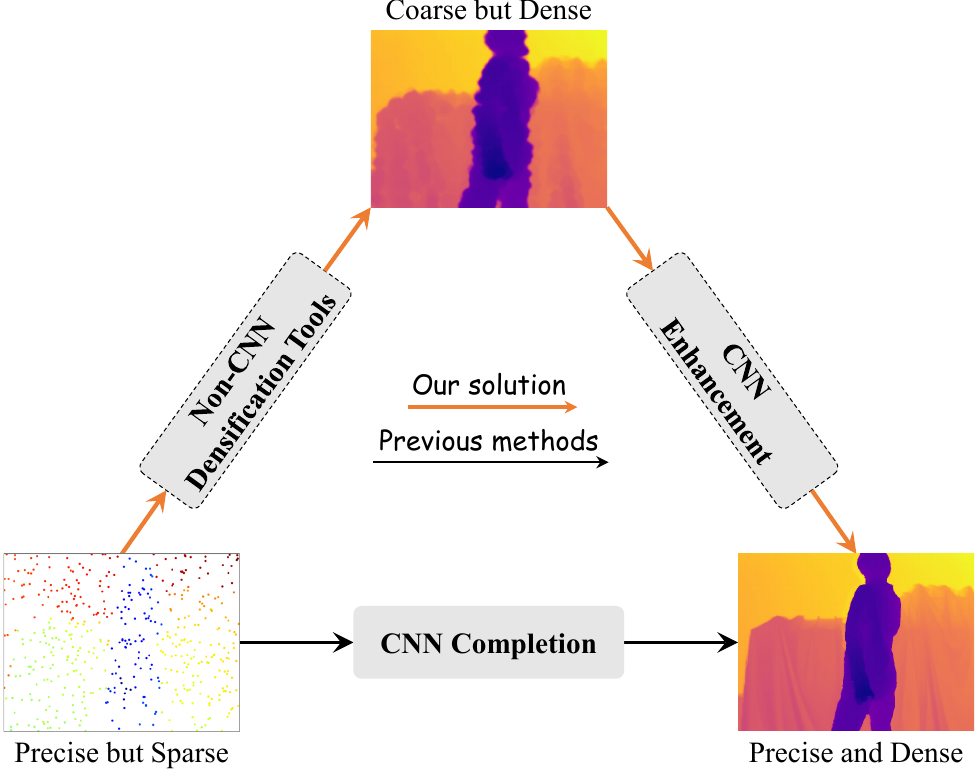}\\
  \caption{Illustration of our main concept. It redefines depth completion as depth enhancement. We first transform sparse depth input via non-CNN densification tools \cite{ku2018ipbasic,levin2004colorization}, yielding coarse but dense depth. Then we predict the precise and dense depth from the coarse depth by leveraging degradation assumption \cite{zhong2023guided,wang2021unsupervised,zhang2021designing}.}\label{main_idea}
\end{figure}

The key challenges of this task lie in the sparse depth input data \cite{Uhrig2017THREEDV,liu2021learning,yan2023desnet,yan2024tri,tang2024bilateral}. We summarize these challenges as mismatch and ambiguity. As shown in Fig.~\ref{main_idea}, the mismatch is caused by the irregularly sampled sparse depth with varying point distributions, while convolutions are spatially invariant with fixed kernel shapes. Moreover, it is ambiguous for convolutions to distinguish between valid and invalid depth measurements since unknown pixels in the sparse depth are recognized as zero.

To address these challenges, several excellent methods \cite{Uhrig2017THREEDV,liu2021learning,tang2024bilateral} have been proposed. For instance, SparseConvs \cite{Uhrig2017THREEDV} proposes sparsity invariant CNNs that explicitly model the location of missing data. To avoid directly convolving on sparse depth, \citet{liu2021learning} employs kernels learned by differentiable kernel regression layers to interpolate sparse data. BPNet \cite{tang2024bilateral} develops bilateral propagation, where the coefficients are generated based on both radiometric difference and spatial distance. These data-driven schemes are robust and impressive. However, they still cannot fully avoid the application of convolution on incomplete depth data, even though the depth data has become denser. 

In this paper, our objective is to develop a novel solution that not only completely resolves the issues of mismatch and ambiguity but also provides a fresh perspective on revisiting the traditional depth completion task. 

To this end, we propose a novel framework called the Selective Image Guided Network (SigNet). As shown in Fig.~\ref{main_idea}, SigNet redefines depth completion as depth enhancement \footnote{Specifically depth super-resolution (DSR). Mainstream DSR methods restore target depth directly from upsampled depth. Consequently, the processes and the end goals of depth completion and DSR are identical, \textit{i.e.}, predicting clear depth from the same-resolution coarse depth.}. It first leverages non-CNN densification tools \cite{ku2018ipbasic,levin2004colorization} to obtain coarse but dense depth. This coarse depth is then enhanced by a CNN-based subnetwork. Inspired by previous approaches \cite{zhong2023guided,zhang2021designing}, we assume that there exists unknown degradation $\mathbf{h}$ between the coarse depth $\mathbf{Z}$ and the target depth $\mathbf{Y}$, described as
\begin{equation}\label{eq_degradation}
\mathbf{Z}=\mathbf{h}\mathbf{Y} + \mathbf{n},
\end{equation}
where $\mathbf{n}$ denotes additive noise. 
Therefore, we can establish a connection between the source depth and the target depth through the degradation. 
A fundamental observation is that degradation typically occurs near edges, suggesting that this phenomenon can be leveraged to enhance RGB-D fusion. This is particularly relevant given that the coarse depth often lacks significant edge details, while the RGB image contains abundant high-frequency components. Specifically, SigNet introduces a Degradation-Aware Decomposition and Fusion (DADF) module. DADF begins by decomposing the implicit degradation $\mathbf{h}$ into multiple patches and estimating their coefficients in the discrete cosine domain. Each patch is then used to adaptively select the most representative high-frequency information from the RGB image. To achieve effective RGB-D aggregation, DADF incorporates a conditional Mamba to encode long-range correlations. Unlike the vanilla Vision Mamba \cite{liu2024vmamba}, which calculates state parameters from the input, DADF uses the degradation patches as additional conditions to generate the state parameters. This approach enables further global high-frequency information interaction. 

In summary, our contributions are as follows:
\begin{itemize}
    \item We introduce a novel framework called SigNet, which redefines depth completion as depth enhancement for the first time. By adopting a degradation-aware approach, SigNet not only offers a new perspective to the field but also fully eliminates the traditional challenges such as mismatch and ambiguity. 
    \item We propose a Degradation-Aware Decomposition and Fusion (DADF) module. Utilizing degradation decomposition, DADF adaptively selects high-frequency components of RGB data to compensate for coarse depth. Furthermore, DADF achieves effective RGB-D fusion by incorporating conditional Mamba, where the state parameters are generated from degradation patches. 
        \item We evaluate SigNet through comprehensive experiments, demonstrating its superiority over previous SOTAs. 
\end{itemize}

\section{Related Work}
\textbf{Depth Completion.} 
Depth completion can be broadly categorized into depth-only and image-guided categories. 

Depth-only methods \cite{Uhrig2017THREEDV,ku2018ipbasic,2018Sparse, 2020Confidence,ma2018self,vangansbeke2019,wang2023lrru,yan2024tri} obtain dense depth directly from a single sparse depth input. For example, IP-Basic \cite{ku2018ipbasic} designs reasonable combinations of basic image processing operations to densify sparse depth without using CNNs. \citet{ma2018self} employ a UNet-like architecture to progressively densify the sparse depth input. FusionNet \cite{vangansbeke2019} designs a lightweight completion network that aggregates both global and local information with confidence adjustment. Moreover, \citet{2020FromLu} introduce a color image branch as auxiliary supervision during training.

Despite advancements in single-modal methods, their performance is often limited without supplementary reference information. To address this challenge, numerous image-guided techniques have been proposed \cite{2020Denseyang,tang2020learning,li2020multi,hu2020PENet, liu2021fcfr,lin2022dynamic,zhang2023cf,zhou2023bev,wang2024improving,tang2024bilateral,yan2024tri}. For example, GuideNet \cite{tang2020learning} and RigNet \cite{yan2022rignet} utilize color images to generate adaptive filtering kernels for depth feature extraction. 
GFormer \cite{rho2022guideformer} and CFormer \cite{zhang2023cf} integrate convolution and transformer mechanisms to capture both local and global representations. LRRU \cite{wang2023lrru} introduces a dynamic kernel scope that transitions from large to small to capture varying dependencies. FuseNet \cite{chen2019learning} and PointDC \cite{yu2023aggregating} incorporate LiDAR point cloud branches to model 3D geometries. 
BEV@DC \cite{zhou2023bev} uses bird’s-eye view to build a point-voxel architecture. Additionally, DepthPrompting \cite{park2024depth} embeds priors of large depth models for better sensor-agnostic generalization. 

 \begin{figure*}[t]
  \centering \includegraphics[width=1.78\columnwidth]{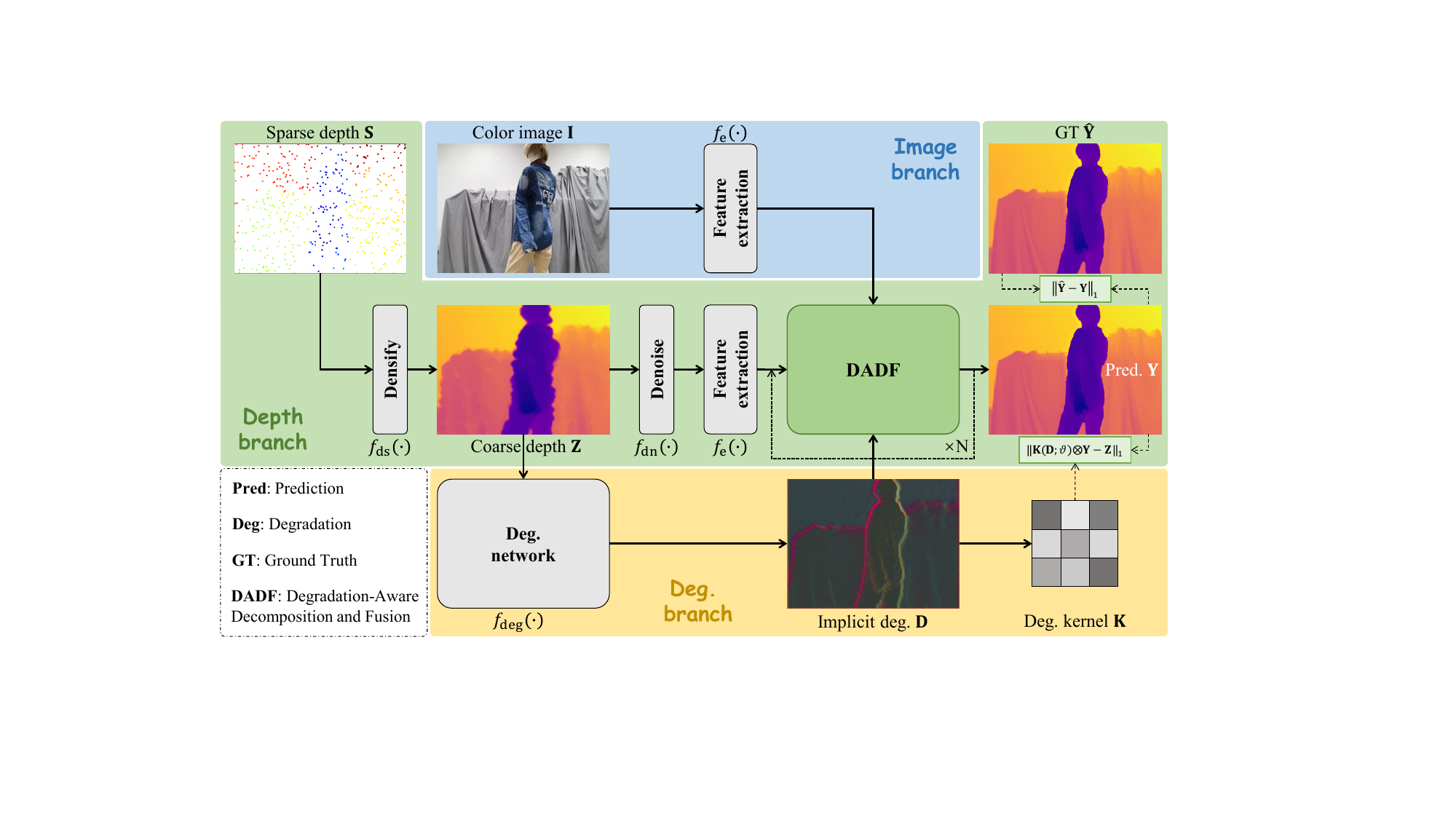}\\
  \vspace{-4pt}
  \caption{Pipeline of SigNet. The sparse depth data is initially filled to create a coarse depth map. We then utilize the degradation assumption in \cref{eq_degradation} to establish a connection between this coarse map and the target prediction. The color image features and degradation representations are employed to guide the multi-modal fusion through our proposed DADF module, as depicted in Fig.~\ref{fig_dadf}.}\label{fig_pipeline}
  \vspace{-4pt}
\end{figure*}

To further advance completion models, spatial propagation networks (SPNs) have been introduced. The initial SPN \cite{liu2017SPN} is designed to learn a pairwise similarity matrix. CSPN \cite{cheng2018depth} applies this technique to depth completion by conducting recursive convolutions with fixed local neighborhood kernels, while CSPN++ \cite{Cheng2020CSPN++} adapts by learning variable kernel sizes. PENet \cite{hu2020PENet} expands the receptive fields using dilated convolutions. Furthermore, NLSPN \cite{park2020nonlocal} integrates non-local neighbors via deformable convolutions. DySPN \cite{lin2023dyspn} generates dynamic non-linear neighbors using an attention mechanism. Most recently, 3D SPNs \cite{cheng2019learning} have been developed to encode 3D geometry. For example, GraphCSPN \cite{liu2022graphcspn} employs additional geometric constraints to regulate 3D propagation. TPVD \cite{yan2024tri} introduces a geometric SPN that constructs the affinity in 2D tri-perspective view spaces and their combined 3D projection space.

\vspace{0.5mm}
\noindent \textbf{Degradation Learning.} 
Recently, color image restoration has been significantly advanced by improvements in degradation representations \cite{yin2022conditional,zhang2021designing,wang2021unsupervised}. For example, \citet{zhang2021designing} construct a sophisticated yet practical degradation model to simulate real-world image degradation for blind image super-resolution. 
Similarly, \citet{zhou2023learning} propose to estimate degradation for each local image region using a degradation-adaptive regression filter for blind super-resolution. \citet{liang2022efficient} present a degradation-adaptive model that specifies the network parameters by estimating the degradation. 
Furthermore, several image restoration methods \cite{zhang2023all,li2022all,zhang2023ingredient} have been proposed to simultaneously address multiple forms of degradation. 
For example, \citet{zhang2023all} develop a multi-degradation image restoration network that incrementally learns degradation representations via clustering, without relying on prior knowledge of the degradation information. 
\citet{zhang2023ingredient} build an ingredient-oriented degradation reformulation network to model the relationship among different images. Additionally, \citet{yao2024neural} introduce a neural degradation representation learning method that captures latent degradation features, adaptively decomposing various types of degradation to enable comprehensive restoration. 

\section{Method}
\textbf{Overview.} 
Fig.~\ref{fig_pipeline} shows our SigNet pipeline that consists of the depth branch, degradation branch, and image branch. In the depth branch, the sparse depth $\mathbf{S}$ is first filled via non-CNN densification tools \cite{ku2018ipbasic,levin2004colorization}, obtaining a coarse depth map $\mathbf{Z}$. According to \cref{eq_degradation}, we employ denoising operators to weaken the negative impact of the noise $\mathbf{n}$, yielding $\mathbf{\hat{Z}}$. Then, we use residual groups \cite{zhang2018image} to extract features of the denoised depth $\mathbf{\hat{Z}}$ and the color image $\mathbf{I}$ in the image branch. In the degradation branch, the implicit degradation $\mathbf{D}$ is generated from a degradation network, where convolutions and residual groups are used. We thus utilize the image feature and degradation $\mathbf{D}$ to facilitate RGB-D fusion via our proposed Degradation-Aware Decomposition and Fusion (DADF). In summary, our SigNet redefines the depth completion as depth enhancement through the degradation assumption in \cref{eq_degradation}, making the model free from the challenges of mismatch and ambiguity.

\subsection{Task Switching}
Degradation learning is widely applied in low-level image restoration \cite{yin2022conditional,zhang2021designing,wang2021unsupervised}, yielding impressive results. These tasks exhibit two key characteristics: 1) both input and target data are typically dense, and 2) the input data generally have a lower resolution than the target data. Actually, the low-resolution input data are typically upsampled using interpolation methods, such as bicubic interpolation. Consequently, the degradation matrix usually consists of a convolution whose kernel is derived from degradation representations and a downsampling function. In contrast, the depth completion task takes sparse depth data as input, and the resolutions of the input and target data are the same. Thus, we reframe depth completion as depth enhancement by drawing an analogy to image restoration.

 \begin{figure}[t]
  \centering
  \includegraphics[width=1\columnwidth]{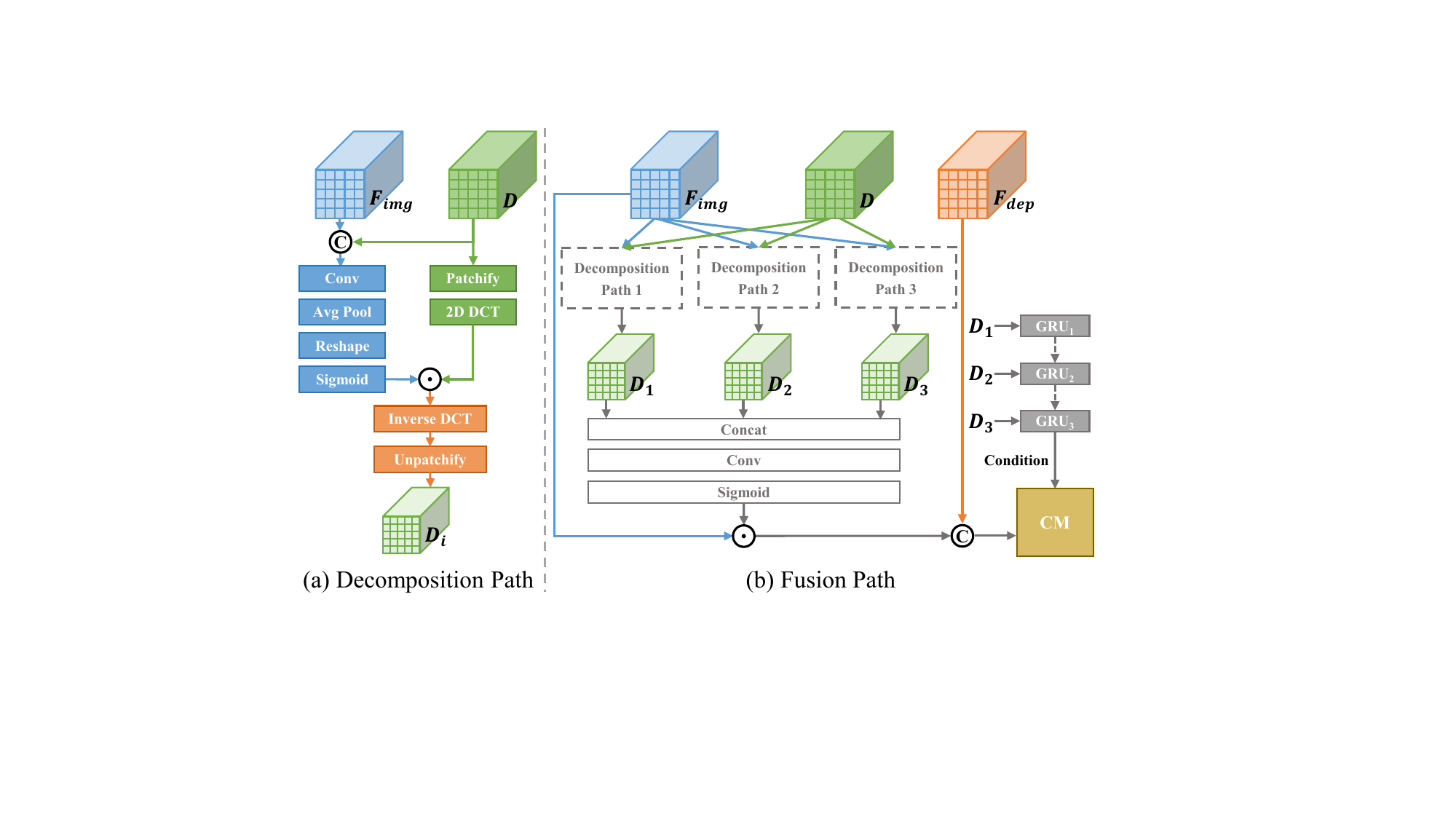}\\
  \vspace{-2pt}
  \caption{Overview of our DADF. $\odot $ and CM refer to element-wise multiplication and the conditional Mamba, respectively.}\label{fig_dadf}
  \vspace{0pt}
\end{figure}

As shown in Fig.~\ref{fig_pipeline}, we first fill the sparse depth $\mathbf{S}$ using non-CNN densification tools \cite{levin2004colorization,ku2018ipbasic}, denoted as $f_{ds}(\cdot)$, resulting in a coarse yet dense depth $\mathbf{Z}$, formulated as: 
\begin{equation}\label{eq_densify}
\mathbf{Z}=f_{ds}(\mathbf{S}).
\end{equation}
As discussed in Sec.~\ref{sec:intro}, this approach eliminates the mismatch and ambiguity caused by direct convolution over irregularly sampled sparse data. Subsequently, we employ a degradation network to estimate the implicit degradation $\mathbf{D}$. The degradation kernel $\mathbf{K}$ is produced from $\mathbf{D}$ using convolutions. Consequently, given the coarse depth $\mathbf{Z}$ and target depth $\mathbf{Y}$, \cref{eq_degradation} can be rewritten based on the task switching as follows:
\begin{equation}\label{eq_z}
\mathbf{Z}=\mathbf{K}(\mathbf{D;\vartheta}) \otimes \mathbf{Y} + \mathbf{n},
\end{equation}
where $\otimes$ indicates convolution operation, $\vartheta$ is the network parameter. Since the resolutions are the same, our degradation equation does not need the downsampling function.

\subsection{Degradation-Aware Decomposition and Fusion}
As we know, degradation typically occurs near edges, suggesting that this cue can be leveraged to enhance RGB-D fusion, especially considering that the coarse depth often lacks significant edge details, while the RGB image contains abundant high-frequency components. Thus, we propose DADF that is shown in Fig.~\ref{fig_dadf}. We denote the image and depth features as \( \mathbf{F}_{img} \) and \( \mathbf{F}_{dep} \), obtaining:
\begin{align}
    &\mathbf{F}_{img}=f_e(\mathbf{I}),\\
    &\mathbf{F}_{dep}=f_e(f_{dn}(\mathbf{Z})),
\end{align}
where $f_e(\cdot)$ and $f_{dn}(\cdot)$ refer to the feature extraction network and denoising function, respectively. 
As presented in Fig.~\ref{fig_dadf}, DADF takes as input the image feature $\mathbf{F}_{img}$, depth feature $\mathbf{F}_{dep}$, and degradation $\mathbf{F}_{D}$.

\begin{algorithm}[t]
\DontPrintSemicolon
  \SetAlgoLined
  \small
      \KwIn {$\mathbf{F}_{rgbd}, \{\mathbf{D}_{i}|i=1,2,3\}:\rm{\textcolor[RGB]{0, 100, 0}{(HW, C)}}$}
      \KwOut {$\hat{\mathbf{F}}_{rgbd}:\rm{\textcolor[RGB]{0, 100, 0}{(HW, C)}}$}
       $\mathbf{A}:{\rm{\textcolor[RGB]{0, 100, 0}{(C, N)}}}\gets \mathbf{Parameter}_{A}$ \;
        $\mathbf{B}:{\rm{\textcolor[RGB]{0, 100, 0}{(HW, N)}}}\gets \mathbf{Linear}_{B}(f_{gru}(\mathbf{F}_{rgbd}, \{\mathbf{D}_{i}|i=1,2,3\}))$  \;
        
        $\mathbf{C}:{\rm{\textcolor[RGB]{0, 100, 0}{(HW, N)}}}\gets \mathbf{Linear}_{C}(f_{gru}(\mathbf{F}_{rgbd}, \{\mathbf{D}_{i}|i=1,2,3\}))$  \;

        $\mathbf{\Delta}:{\rm{\textcolor[RGB]{0, 100, 0}{(HW, C)}}}\gets {\rm{log}(1+{\rm{exp}}}(\mathbf{Linear}_{\Delta}(f_{gru}(\mathbf{F}_{rgbd}, \{\mathbf{D}_{i}|i=1,2,3\}))+\mathbf{Parameter}_{\Delta}))$  \;

        $\mathbf{\overline{A}}:{\rm{\textcolor[RGB]{0, 100, 0}{(HW, C, N)}}}\gets \rm{exp(\mathbf{\Delta}\otimes \mathbf{A})}$  \;
        
		$\mathbf{\overline{B}}:\rm{\textcolor[RGB]{0, 100, 0}{(HW, C, N)}}\gets \mathbf{\Delta}\otimes \mathbf{B}$  \;
        
		$\hat{\mathbf{F}}_{rgbd} \gets {\rm{SSM}(\mathbf{\overline{A}}, \mathbf{\overline{B}}, \mathbf{C})}(\mathbf{F}_{rgbd})$  \;
        \Return $\hat{\mathbf{F}}_{rgbd}$ \;
  \caption{Conditional SSM Block}
  \label{algor}
\end{algorithm}

\noindent \textbf{Decomposition.} 
To extract the most representative components of $\mathbf{F}_{D}$, we first apply the discrete cosine transform (DCT) to convert it into the frequency domain. Next, it adaptively learns a spectrum mask to enhance informative frequency bands while suppressing irrelevant ones. Finally, we reconstruct the spatial representation using the inverse DCT. This process is formulated as:
\begin{equation}\label{eq_dct}
\{\mathbf{D}_{i}|i=1,2,3\}=f_{dct}(\mathbf{D}).
\end{equation}
We apply concatenation, convolution, pooling, and Sigmoid operations, collectively denoted as $f_m(\cdot)$, to generate a mask $m$. This mask is then multiplied with the combined frequency components to produce an updated mask $\hat{m}$, enabling the adaptive selection of the representative image feature $\hat{\mathbf{F}}_{img}$. The process is defined as:
\begin{align}
    &m=f_m(\mathbf{F}_{img}, \mathbf{D}),\\
    &\hat{m}=f_{idct}(m \cdot (f_{ac}(\sum\nolimits_{i=1}^{3}{{{\mathbf{D}}_{i}}}))),\\
    & \hat{\mathbf{F}}_{img}=\hat{m} \cdot \mathbf{F}_{img},
\end{align}
where $f_{idct}$ refers to inverse DCT and $f_{ac}(\cdot)$ represents the combination function of addition and convolutions.

\begin{table*}[t]
\small
\centering
\renewcommand\arraystretch{1.02}
\resizebox{0.92\textwidth}{!}{
\begin{tabular}{c|c|c|ccccc|c}
\toprule
Dataset   & Method    & Params. (M) $\downarrow$      & RMSE (mm) $\downarrow$      & REL $\downarrow$     &${\delta }_{1} (\%)$  $\uparrow$ & ${\delta }_{{2}} (\%)$ $\uparrow$ & ${\delta }_{{3}} (\%)$  $\uparrow$  & Venue \\ 
\midrule 
\multirow{9}{*}{NYUv2} 
& Bilateral \cite{buades2005non}         & - & 532    & 0.132   & 85.1  & 93.5  & 95.9 & CVPR 2005 \\
& S2D \cite{ma2018self}       & 26.1 & 230    & 0.054   & 94.5  & 97.3  & 98.9 & ICRA 2019 \\
& CSPN \cite{2018Learning} & \underline{17.4} & 173    & 0.020   & 96.3  & 98.6  & 99.5 & ECCV 2018 \\
& DfuseNet \cite{shivakumar2019dfusenet}    & - & 156    & 0.016   & 98.8  & \underline{99.7}  & \underline{99.9} & ITSC 2019\\ 
& DM-LRN \cite{senushkin2021decoder}      & - & 205    & 0.014   & 98.8  & 99.6  & \underline{99.9} & IROS 2021 \\
& RDF-GAN \cite{wang2022rgb}       & - & 139    & 0.013   & 98.7  & 99.6  & \underline{99.9} & CVPR 2022 \\
& AGG-Net \cite{chen2023agg}      & 129.1 & 92    & 0.014   & 99.4  & \textbf{99.9}  & \textbf{100.0}  & ICCV 2023 \\
& BPNet \cite{tang2024bilateral}         & 89.9  & 88  & \textbf{0.011}  & 99.6  & \textbf{99.9}  & \textbf{100.0}  & CVPR 2024 \\
& \textbf{SigNet (ours)} & \textbf{3.3}  & \textbf{83}  & \underline{0.012}  & \textbf{99.7}  & \textbf{99.9}  & \textbf{100.0}  & CVPR 2025 \\
\midrule
\multirow{7}{*}{DIML} 
& Bilateral \cite{buades2005non}         & - & 636    & 0.189   & 83.0  & 88.8  & 92.4 & CVPR 2005 \\
& CSPN \cite{2018Learning} & \underline{17.4} & 162    & 0.033   & 96.1  & 98.7  & 99.6 & ECCV 2018 \\
& DfuseNet \cite{shivakumar2019dfusenet}    & - & 143    & 0.023   & 98.4  & 99.4  & \underline{99.9} & ITSC 2019\\ 
& DM-LRN \cite{senushkin2021decoder}      & - & 149    & 0.015   & 99.0  & 99.6  & \underline{99.9} & IROS 2021 \\
& AGG-Net \cite{chen2023agg}      & 129.1 & 78    & \underline{0.011}   & \underline{99.6}  & \textbf{99.9}  & \textbf{100.0}  & ICCV 2023 \\
& BPNet \cite{tang2024bilateral}         & 89.9  & \underline{68}  & \underline{0.011}  & 99.4  & \underline{99.8}  & \textbf{100.0} & CVPR 2024 \\
& \textbf{SigNet (ours)} & \textbf{3.3}  & \textbf{55}  & \textbf{0.006}  & \textbf{99.7}  & \textbf{99.9}  & \textbf{100.0} & CVPR 2025 \\
\midrule
\multirow{8}{*}{SUN RGBD} 
& S2D \cite{ma2018self}       & 26.1 & 329    & 0.074   & 93.9  & 97.0  & 98.1 & ICRA 2019 \\
& CSPN \cite{2018Learning} & \underline{17.4} & 295    & 0.137   & 95.6  & 97.5  & 98.4 & ECCV 2018 \\
& DLiDAR \cite{Qiu_2019_CVPR}    & 53.4 & 279    & 0.061   & 96.9  & 98.0  & 98.4 & CVPR 2019 \\ 
& DM-LRN \cite{senushkin2021decoder}      & - & 267    & 0.063   & 97.6  & 98.2  & 98.7 & IROS 2021 \\
& RDF-GAN \cite{wang2022rgb}       & - & 255    & 0.059   & 96.9  & 98.4  & 99.0 & CVPR 2022 \\
& AGG-Net \cite{chen2023agg}      & 129.1 & 152    & 0.038   & 98.5  & \underline{99.0}  & \underline{99.4}  & ICCV 2023 \\
& BPNet \cite{tang2024bilateral}         & 89.9  & \underline{124} & \underline{0.025}  & \underline{98.9}  & \textbf{99.4}  & \textbf{99.8}  & CVPR 2024 \\
& \textbf{SigNet (ours)} & \textbf{3.3}  & \textbf{116}  & \textbf{0.022} & \textbf{99.0}  &  \textbf{99.4}  & \textbf{99.8} & CVPR 2025 \\
\midrule
\multirow{10}{*}{TOFDC} 
& CSPN \cite{2018Learning}          & \underline{17.4}   & 224  & 0.042  & 94.5  & 95.3  & 96.5  & ECCV 2018 \\
& GuideNet \cite{tang2020learning}  & 73.5   & 146  & 0.030  & 97.6  & 98.9  & 99.5  & TIP 2020 \\
& PENet \cite{hu2020PENet}          & 131.5  & 241  & 0.043  & 94.6  & 95.3  & 95.5  & ICRA 2021 \\
& NLSPN \cite{park2020nonlocal}     & 25.8   & 174  & 0.029  & 96.4  & 97.9  & 98.9  & ECCV 2020 \\
& CFormer \cite{zhang2023cf}        & 83.5   & 113  & 0.029 & \underline{99.1}   & \underline{99.6}  & \textbf{99.9}  & CVPR 2023 \\
& RigNet \cite{yan2022rignet}       & 65.2   & 133  & 0.025  & 97.6  & 99.1  & \underline{99.7}  & ECCV 2022 \\
& GraphCSPN \cite{liu2022graphcspn} & 26.4   & 253  & 0.052  & 92.0  & 96.9  & 98.7  & ECCV 2022 \\ 
& PointDC \cite{yu2023aggregating}  & 25.1   & 109  & 0.021  & 98.5  & 99.2  & 99.6  & ICCV 2023 \\
& TPVD \cite{yan2024tri}            & 31.2   & \underline{92}    & \underline{0.014}   & \underline{99.1}  & \underline{99.6}  & \textbf{99.9}  & CVPR 2024 \\ 
& \textbf{SigNet (ours)} & \textbf{3.3}  & \textbf{87}  & \textbf{0.012}  & \textbf{99.2}  & \textbf{99.7}  & \textbf{99.9} & CVPR 2025 \\
\bottomrule
\end{tabular}
}
\vspace{-5pt}
\caption{Quantitative comparison on four depth completion datasets. The \textbf{best} and \underline{suboptimal} results are highlighted. Note that most of the results on NYUv2, DIML, and SUN RGBD are borrowed directly from AGG-Net \cite{chen2023agg}, while those of TOFDC are from TPVD \cite{yan2024tri}.}\label{tab_nyu_diml}
\end{table*}

\noindent \textbf{Fusion.} 
As a result, we further leverage the decomposed degradation $\mathbf{D}_{i}$ to enable effective RGB-D fusion. It is well-known that Mamba \cite{gu2023mamba} is good at handling long-range correlations. Inspired by recent Vision Mamba \cite{liu2024vmamba}, we present a variant, \emph{i.e.}, Conditional Mamba (CM). As shown in Fig.~\ref{fig_dadf}, the selected image feature $\hat{\mathbf{F}}_{img}$ and raw depth feature $\mathbf{F}_{dep}$ are concatenated into CM, where the decomposed degradation $\mathbf{D}_{i}$ serves as the condition:
\begin{align}
&\mathbf{F}_{rgbd}=f_{cc}(\hat{\mathbf{F}}_{img}, \mathbf{F}_{dep}),\\
&\hat{\mathbf{F}}_{rgbd}=f_{cm}(\mathbf{F}_{rgbd}, \{\mathbf{D}_{i}|i=1,2,3\}),
\end{align}
where $f_{cc}$ denotes the combination function of concatenation and convolutions, $f_{cm}$ is the Condition Mamba function. As reported in Alg.~\ref{algor}, the state parameters $\mathbf{B}$ and $\mathbf{C}$ are calculated based on both the RGB-D input and degradation using gated recurrent convolution units.

\subsection{Loss Function}
As shown in our pipeline, the total loss function $\mathcal L_{t}$ consists of a reconstruction loss $\mathcal L_{r}$ and a self-supervised degradation learning loss $\mathcal L_{d}$. 
Let $\mathbf{\hat{Y}}$ and $\mathbf{Y}$ denote the ground truth depth and the predicted depth, respectively. The reconstruction loss $\mathcal L_{r}$ is defined as:
\begin{equation}\label{eq:Lcont1}
   \mathcal L_{r}=\frac{1}{N}\sum_{i=1}^{N}  ||\mathbf{\hat{Y}}_{i} -\mathbf{Y}_{i}  ||_{1}  ,
\end{equation}
where $N$ represents the number of training samples, and $|| \cdot ||_{1} $ denotes the $L_{1}$ norm. 
Subsequently, we introduce the degradation loss $\mathcal L_{d}$ to facilitate the degradation learning process through:
 \begin{equation}\label{eq:Lcont2}
   \mathcal L_{d}=\frac{1}{N}\sum_{i=1}^{N} ||\mathbf{K}\otimes \mathbf{Y}- \mathbf{Z}  ||_{1} ,
\end{equation}
where $\mathbf{K}$ refers to the degradation kernel learned from the implicit degradation representation $\mathbf{D}$. $\mathbf{Z}$ corresponds to the coarse depth. $\otimes$ denotes the convolution operation, where $\mathbf{K}$ serves as the convolution kernel. 
As a result, the total training loss is formulated as: 
 \begin{equation}\label{eq:Lcont3}
   \mathcal L_{t}=\mathcal L_{r} + \lambda \mathcal L_{d},
\end{equation}
where $\lambda$ is a hyper-parameter, empirically set to $0.1$. By the constraint of the auxiliary loss $\mathcal L_{d}$, our model can build additional connections between the input and output, thus predicting more accurate depth results. 

\begin{figure*}[t]
\centering
\includegraphics[width=1.86\columnwidth]{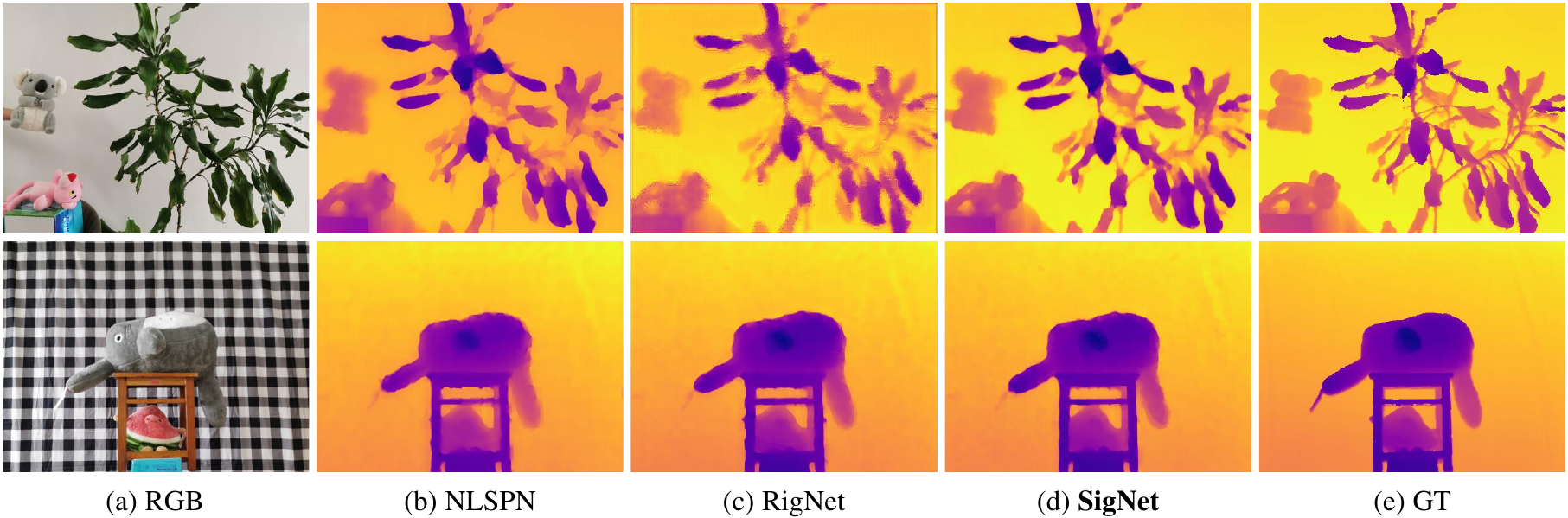}\\
\vspace{-2pt}
\caption{Visual comparison with SOTA methods on TOFDC dataset, including NLSPN \cite{park2020nonlocal}, RigNet \cite{yan2022rignet}, and our SigNet.}\label{fig_tofdc}
\end{figure*}

\begin{figure*}[t]
\centering
\includegraphics[width=1.86\columnwidth]{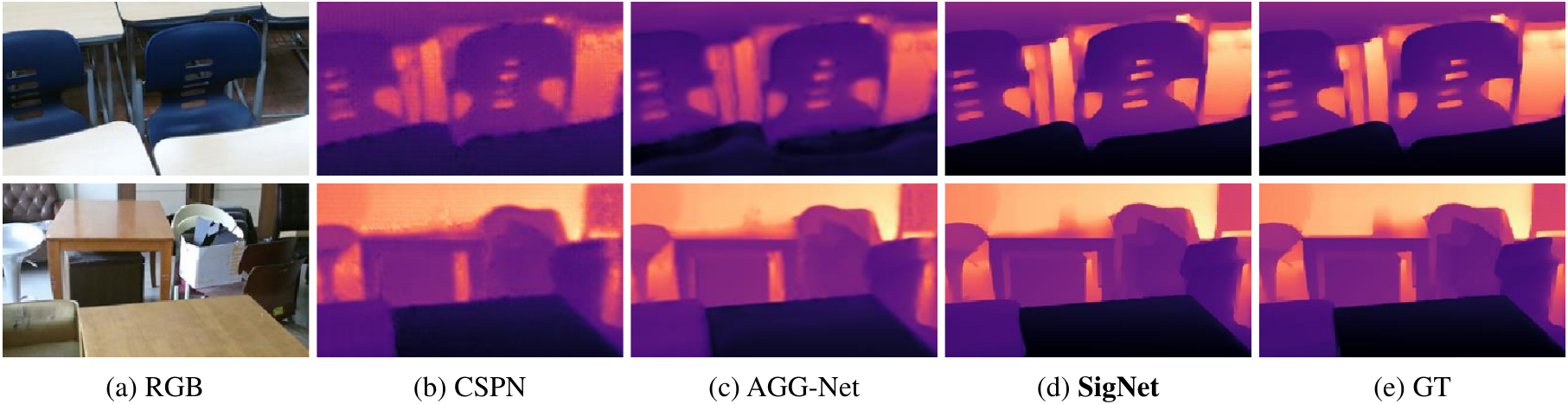}\\
\vspace{-2pt}
\caption{Visual comparison with SOTA approaches on DIML dataset, including CSPN \cite{2018Learning}, AGG-Net \cite{chen2023agg}, and our SigNet.}\label{fig_diml}
\end{figure*}

 \begin{figure*}[!ht]
  \centering
  \includegraphics[width=1.86\columnwidth]{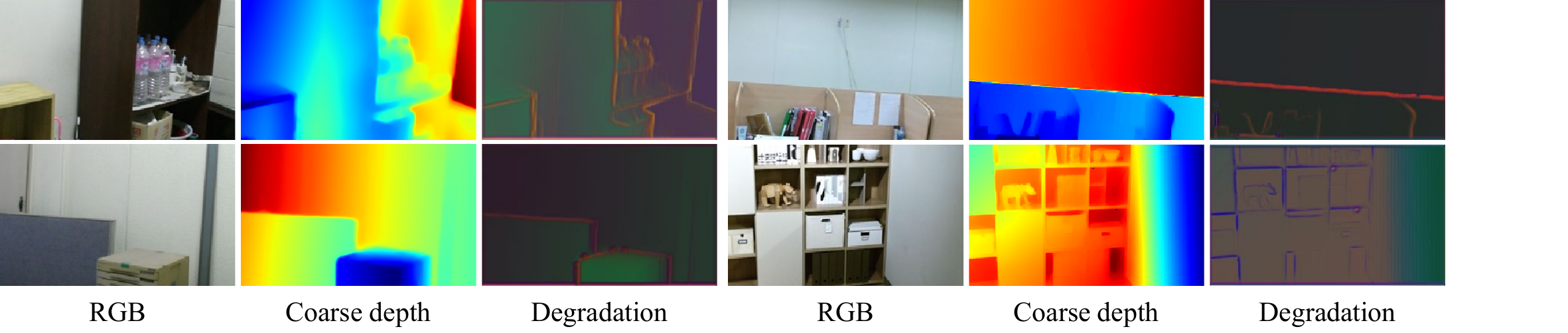}\\
  \vspace{-2pt}
  \caption{Visual examples of filled coarse depth maps and degradation representations on the DIML dataset.}\label{fig_deg_example}
\end{figure*}

\section{Experiment}
We evaluate our SigNet method on four popular datasets, including NYUv2, DIML, SUN RGBD, and TOFDC benchmarks. For NYUv2, DIML, and SUN RGBD, we maintain the settings used in AGG-Net \cite{chen2023agg}, while for TOFDC, we follow the settings of TPVD \cite{yan2024tri}. Refer to our supplementary material for additional details on the experimental setups.

\subsection{Comparison with SOTA}
To assess the effectiveness of our SigNet, we benchmark it against the traditional bilateral filtering method \cite{buades2005non} and some recent SOTA deep learning approaches such as \cite{ma2018self,shivakumar2019dfusenet,park2020nonlocal,chen2023agg,tang2024bilateral}. The error and accuracy results for the competing models are sourced from their original publications or obtained by retraining them following the default settings.

\subsubsection{Performance Analysis}
\noindent \textbf{On NYUv2.} As a widely used benchmark dataset, NYUv2 is often employed to evaluate the effectiveness of models. From the first row of Table~\ref{tab_nyu_diml}, we can observe that SigNet achieves competitive performance on the NYUv2 dataset, with an RMSE that is $5mm$ lower than the second-best method. Additionally, although our method lags behind BPNet \cite{tang2024bilateral} by only $0.001$ in the REL metric, SigNet significantly reduces the number of parameters by $96\%$. 

\noindent \textbf{On DIML.} DIML is collected from both indoor and outdoor scenes, suffering from severe edge shadows and irregular holes. Our method is implemented using the same experimental settings as previous approaches \cite{senushkin2021decoder, chen2023agg}. As shown in the second row of Table~\ref{tab_nyu_diml}, SigNet demonstrates the best performance across all evaluation metrics. For example, compared to the suboptimal model, our approach decreases the RMSE by $19\%$ and the REL by $45\%$.

\noindent \textbf{On SUN RGBD.} The large-scale SUN RGB-D dataset is leveraged to evaluate the generalization capability of our SigNet across different sensors and scenes. Table~\ref{tab_nyu_diml} (third row) reveals that our method surpasses previous state-of-the-art approaches across all metrics. Notably, SigNet achieves a $8mm$ lower RMSE than the latest BPNet \cite{tang2024bilateral} and a $36mm$ lower RMSE compared to AGG-Net \cite{chen2023agg}.

\noindent \textbf{On TOFDC.} TOFDC encompasses extensive data captured under various lighting conditions and in open spaces. It is utilized to further assess the performance of the proposed model. The quantitative results in the last row of Table~\ref{tab_nyu_diml} indicate that our method achieves state-of-the-art performance. Compared to the second-best TPVD \cite{yan2024tri}, SigNet reduces the RMSE by $5mm$ and the REL by $0.002$, while also decreasing the number of parameters by $27.9M$.

\subsubsection{Visualization Analysis}
Figs.~\ref{fig_tofdc} and \ref{fig_diml} showcase visual results on TOFDC and DIML, highlighting SigNet's ability to recover precise depth structures. Compared to previous models, our method yields sharper predictions, such as the leaves in Fig.~\ref{fig_tofdc} and the chair in Fig.~\ref{fig_diml}, achieving closer alignment with the ground truth. These results affirm the effectiveness of our approach in enhancing depth completion. Additionally, Fig.~\ref{fig_deg_example} presents some degradation examples, further demonstrating the ability of SigNet to model and address such challenges.

\begin{table}[t]
\centering
\footnotesize
\renewcommand\arraystretch{1.02}
\resizebox{0.45\textwidth}{!}{
\begin{tabular}{l|ccccc}
\toprule
Density          & 1\%    & 5\%    & 35\%   & 65\%  & 95\%  \\
\midrule
TPVD \cite{yan2024tri}  & 156.6  & 139.7  & 112.2  & 97.3  & 92.8  \\
\textbf{SigNet}  & \textbf{144.6}  & \textbf{121.5}  & \textbf{99.4}   & \textbf{90.1}  & \textbf{84.3}  \\
\textit{Improv.} $\uparrow$ & 12.0 & 18.2 & 12.8 & 7.2 & 8.5 \\
\bottomrule
\end{tabular}
}
\vspace{-2pt}
\caption{Density comparison of TPVD and SigNet on TOFDC.}
\label{tab_valid_points}
\end{table}

 \begin{figure}[t]
  \centering
  \includegraphics[width=0.943\columnwidth]{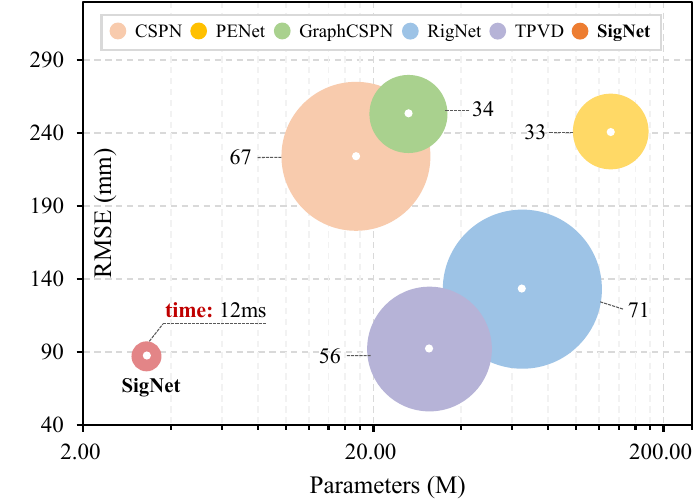}\\
  \vspace{-2pt}
  \caption{Complexity comparison on the TOFDC dataset, where larger circle areas indicate higher inference time.}\label{fig_params}
  \vspace{-2pt}
\end{figure}

\subsubsection{Number of Valid Points}
Tab.~\ref{tab_valid_points} lists a comparison of different valid points on DIML. It indicates that our SigNet consistently outperforms TPVD, achieving an average improvement of 9.7\% across the five densities. These results highlight the generalization capability of our method across varying valid points. 

\subsubsection{Complexity Analysis}
Fig.~\ref{fig_params} shows that our method achieves an excellent balance among RMSE, parameters, and inference time. For example, compared to the fewer parameter CSPN \cite{2018Learning}, our SigNet not only reduces parameters by $81\%$ and inference time by $82\%$, but also decreases RMSE by $61\%$. Additionally, our method surpasses the latest TPVD \cite{yan2024tri} with a $5\%$ lower RMSE and $78\%$ lower FPS, while reducing parameters by $89\%$ and inference time by $78\%$.

\subsection{Ablation Study}
\noindent \textbf{SigNet Designs.} 
Table~\ref{tab_ablation_on_signet} presents the ablation results on the DIML test split. The iteration of DADF is set to five. 
SigNet-i serves as the baseline, where both the degradation bridge and DADF module are removed, and the RGB-D features are fused using addition. Building on this, SigNet-ii redefines depth completion as depth enhancement through densification and degradation, concatenating the RGB and degradation features, resulting in a 13 mm improvement. This highlights the superiority of our task transformation approach. Additionally, SigNet-iii incorporates a Gaussian denoising operator to further reduce error. To better utilize the degradation, SigNet-iv applies a Sigmoid function and element-wise multiplication to obtain an attention map, enabling adaptive selection of RGB features. Consequently, performance improves significantly from 92 mm to 74 mm. Furthermore, SigNet-v employs degradation-aware decomposition, reducing the error by an additional 6 mm. The results of SigNet-iii, iv, and v demonstrate the effectiveness of our adaptive feature selection strategy. Notably, in SigNet-iii, iv, and v, the selected RGB and depth features are encoded through concatenation and convolutions. Finally, with the introduction of our conditional Mamba, SigNet-vi surpasses SigNet-v by 13 mm in RMSE, representing a significant improvement. Overall, each component positively impacts the baseline.

\begin{table}[t]
\Large
\centering
\renewcommand\arraystretch{1.02}
\resizebox{0.478\textwidth}{!}{
\begin{tabular}{l|cc|cccc|c}
\toprule
\multirow{2}{*}{SigNet} & \multicolumn{2}{c|}{Deg. Bridge} & \multicolumn{4}{c|}{RGB-D Fusion} & RMSE \\ \cline{2-7} 
& Densify  & Denoise   & Conc. & Att. &  DADF-D   & DADF-F  & (mm)      \\ 
\midrule
i   &             &             &&             &             && 108            \\
ii   & \checkmark  &            & \checkmark   &        &    && 95             \\
iii  & \checkmark  & \checkmark  & \checkmark  &    &        && 92             \\
iv & \checkmark  & \checkmark  && \checkmark  &              && 74             \\
v  & \checkmark  & \checkmark  &&   & \checkmark             && 68             \\
v  & \checkmark  & \checkmark  &&   & \checkmark  & \checkmark  &  \textbf{55}          \\
\bottomrule
\end{tabular}
}
\caption{Ablation study of our SigNet on the DIML dataset. Conc.: concat, Att.: attention, DADF-D: degradation decomposition in DADF module, DADF-F: conditional Mamba fusion.}
\label{tab_ablation_on_signet}
\end{table}

 \begin{figure}[t]
  \centering
  \includegraphics[width=0.95\columnwidth]{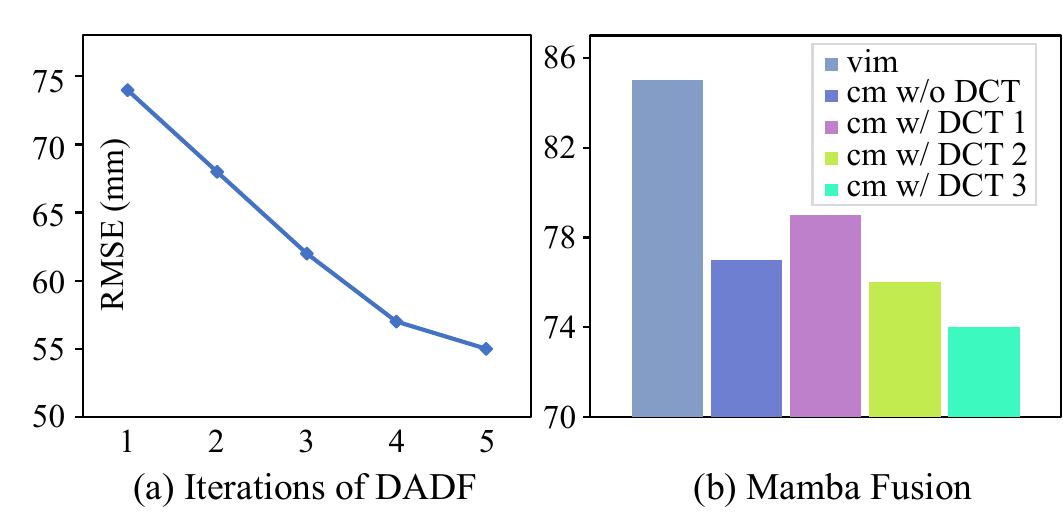}\\
  \caption{Ablation study of our DADF on the DIML dataset.}\label{fig_ablation_dadf}
\end{figure}

\vspace{0.5mm}
\noindent \textbf{Mamba Fusion and Iteration of DADF.} 
Fig.~\ref{fig_ablation_dadf} (a) illustrates the ablation study of Mamba fusion on the DIML dataset. For rapid validation, we employ DADF only once. The baseline ViM directly encodes the combined RGB-D features using the vanilla Vision Mamba \cite{liu2024vmamba}. When we apply our Conditional Mamba with initial degradation and RGB-D features as input, the error decreases by 8 mm. Additionally, the introduction of degradation decomposition using DCT further enhances performance. We also observe that the results improve as the number of DCT components increases, highlighting the effectiveness of our gradual DCT decomposition design. 
Fig.~\ref{fig_ablation_dadf} (b) presents the ablation study of our DADF with different iterations on the DIML dataset. We find that performance improves with an increasing number of iterations. For an optimal balance between efficiency and effectiveness, we set the final iteration count to five. Moreover, as shown in Fig.\ref{fig_dadf}, the degradation $\mathbf{D}$ can be decomposed into $n$ components using DCT. As indicated in Tab.\ref{tab_ablation_path}, the error gradually decreases as $n$ increases. However, considering the computational complexity, we select the top-3 most representative components to focus on the high-frequency components of $\mathbf{F}_{img}$ and update the Mamba condition accordingly.

\begin{table}[t]
\centering
\footnotesize
\renewcommand\arraystretch{1.02}
\resizebox{0.45\textwidth}{!}{
\begin{tabular}{l|ccccc}
\toprule
Path number & $3$  & $5$  & $7$  & $9$  & $11$  \\
\midrule
RMSE (mm)  & 55.0  & 54.6  & 54.5  & 54.1  & \textbf{54.0}  \\
\bottomrule
\end{tabular}
}
\caption{Ablation on the number of decomposition path in DADF.}
\label{tab_ablation_path}
\end{table}

\subsection{Failure Case}
\begin{figure}[t]
  \centering
  \includegraphics[width=1\columnwidth]{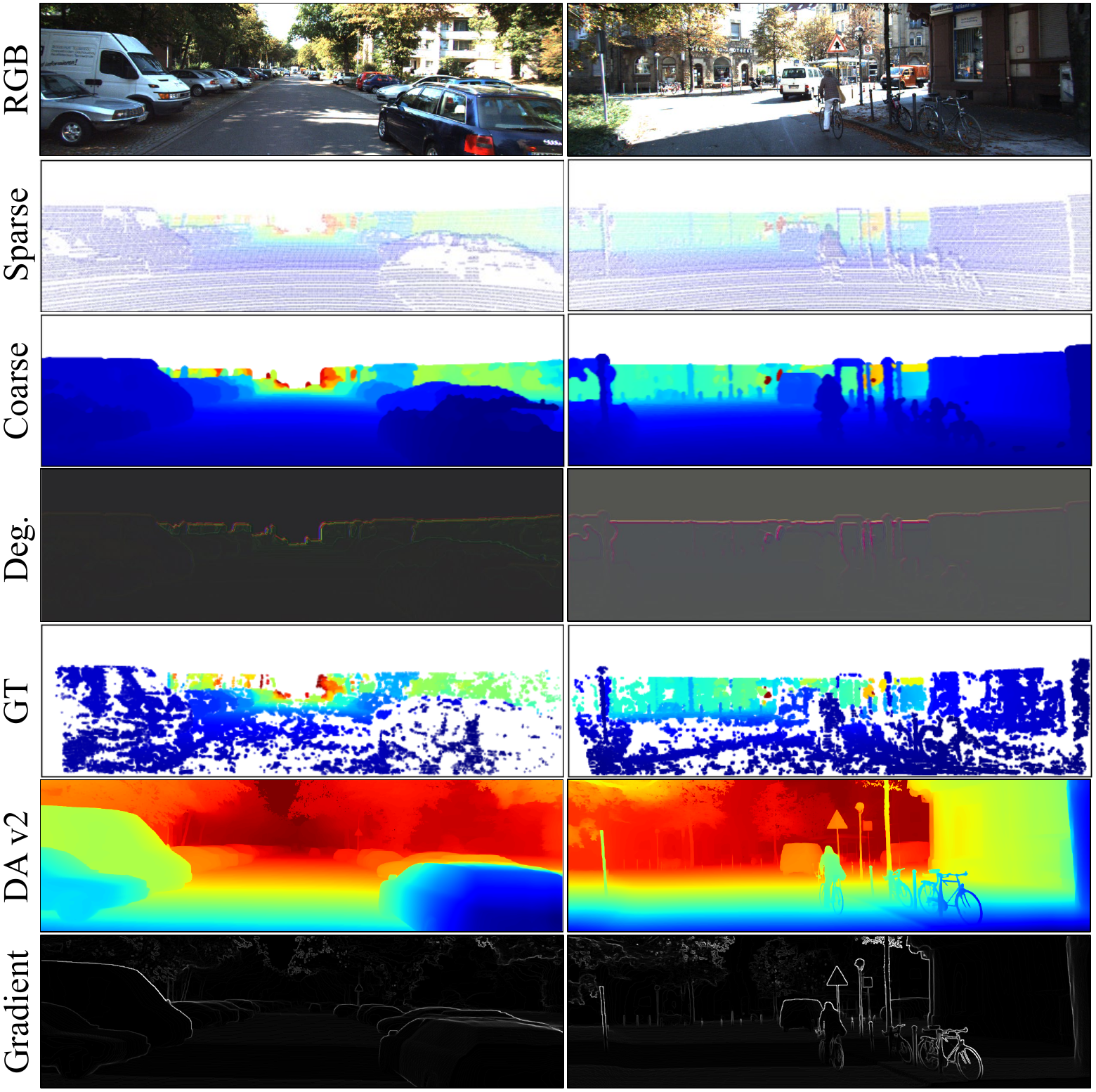}\\
  \vspace{-5pt}
  \caption{Visual results of failure cases on KITTI benchmark \cite{Uhrig2017THREEDV}. DA v2 denotes the large depth model DepthAnything v2 \cite{yang2024depth}.}\label{fig_failure_case}
  \vspace{-6pt}
\end{figure}

Our method is primarily evaluated in indoor scenes and some simple outdoor scenes with closer distances. By constraining dense predictions with coarse depth maps, SigNet can effectively predict their degradation representations. However, as depicted in Fig.~\ref{fig_failure_case}, SigNet fails to estimate reasonable degradation on the outdoor KITTI benchmark \cite{Uhrig2017THREEDV}. We identify two main reasons for this:

\textbf{1)} The ground truth depth maps of KITTI are very sparse, with only about 30\% valid pixels. This sparsity leads to inconsistent depth edges, resulting in a relatively weak constraint. For instance, the degradation in Fig.~\ref{fig_failure_case} even introduces some initial LiDAR lines. 

\textbf{ 2)} The depth distance of KITTI data is much larger, which significantly weakens edge details. As observed, the degradation in Fig.~\ref{fig_failure_case} can only learn very coarse and blurry representations near edges at short range, while at long range, it learns incorrect representations. 

One potential solution is to provide the model with supervisory signals regarding dense contents and precise edges. It is well-known that large depth models can produce dense depth with impressive sharp edges. For example, we employ DepthAnything v2 \cite{yang2024depth} to generate the corresponding dense depth. We discover that the edges of cars and even tree leaves are much clearer. Although the depth scale does not match that of KITTI, we can generate a map that offers constraints near edges via gradient or boundary extraction techniques \cite{wang2024sgnet}. Certainly, the results of DepthAnything v2 \cite{yang2024depth} or DepthAnything v2 itself can also be utilized to facilitate depth completion, warranting further exploration.

\section{Conclusion}
In this paper, we redefined depth completion as depth enhancement for the first time with our novel framework, SigNet. SigNet leveraged non-CNN densification tools to transform sparse depth data into coarse yet dense depth, fully eliminating the mismatch and ambiguity caused by direct convolution over irregularly sampled sparse data. More importantly, by establishing a self-supervised degradation bridge, SigNet effectively fused RGB-D data. It utilized degradation decomposition to dynamically select high-frequency components of RGB and build a conditional Mamba for the depth enhancement. Extensive experiments validated the superiority of SigNet, highlighting its potential for depth related applications.

{
    \small
    \bibliographystyle{ieeenat_fullname}
    \bibliography{main}
}


\setcounter{page}{1}
\maketitlesupplementary

\section{Setup}

\subsection{Dataset}

\noindent \textbf{NYUv2} \cite{silberman2012indoor} is the most commonly used dataset for depth completion, consisting of 1,449 sets from 464 different indoor scenes using Microsoft Kinect. AGG-Net divides this dataset into 420 images for training and 1,029 for testing. The initial resolution of the RGB-D pairs is $640\times 480$, which are randomly cropped and resized to $324\times 288$.

\vspace{1mm}
\noindent \textbf{DIML} \cite{cho2019large} is a newly introduced dataset featuring a series of RGB-D frames captured by the Kinect V2 for indoor scenes and the ZED stereo camera for outdoor scenes. In addition to the usual invalid patterns, it includes numerous edge shadows and irregular holes, making it ideal for evaluating the adaptability of depth completion models to various invalid patterns. AGG-Net focuses solely on the indoor portion of the dataset, which comprises 1,609 RGB-D pairs for training and 503 pairs for testing. The resolution is randomly cropped and resized from $512\times 288$ to $320\times 192$. 

\vspace{1mm}
\noindent \textbf{SUN RGB-D} \cite{song2015sun} is a comprehensive dataset containing 10,335 refined RGB-D pairs, captured using four different sensors across 19 major scene categories. According to the official scheme, the training set includes 4,845 pairs, while the testing set comprises 4,659 pairs. The resolution is randomly cropped and resized from $730\times 530$ to $384\times 288$. 

\vspace{1mm}
\noindent \textbf{TOFDC} \cite{yan2024tri} is collected using the time-of-flight (TOF) depth sensor and RGB camera of a Huawei P30 Pro, encompassing various scenes such as textures, flowers, bodies, and toys under different lighting conditions and in open spaces. It includes 10,000 RGB-D pairs of resolution $512\times 384$ for training and 560 pairs for evaluation. The ground truth depth maps were captured by a Helios TOF camera.

\subsection{Metric}

Following prior works \cite{Qiu_2019_CVPR,park2020nonlocal,tang2020learning,chen2023agg,yan2024tri}, we evaluate our model using the Root Mean Squared Error (RMSE), Absolute Relative Error (REL), and Accuracy under the threshold ${{\delta }_{1.25^i}}$, where $i = 1, 2, 3$. These metrics provide a comprehensive assessment of the model's performance in terms of depth estimation accuracy and error tolerance. The specific definitions and calculations are provided in Tab.~\ref{tab_metric}.

\subsection{Implementation Detail}
Our SigNet is implemented in PyTorch and trained on a single NVIDIA RTX 3090 GPU. We train the model for 20 epochs using the Adam optimizer \cite{Kingma2014Adam} with momentum parameters \(\beta_{1} = 0.9\) and \(\beta_{2} = 0.999\), an initial learning rate of \(1 \times 10^{-4}\), and a weight decay of \(1 \times 10^{-6}\). The balance coefficient \(\gamma\) in the loss function is set to 0.1. To further improve model performance and generalization, we apply data augmentation techniques, including random cropping, flipping, and normalization during training, which helps the model better handle variations in input data.

\begin{table}[t]
\small
\centering
\renewcommand\arraystretch{1.7}
\resizebox{0.44\textwidth}{!}{
\begin{tabular}{ll}
\toprule
\multicolumn{2}{l}{For one pixel $p$ in the valid pixel set $\mathbb{P}$ of $\mathbf{y}$: } \\ 
-- RMSE              & $\sqrt{\frac{1}{|\mathbb{P}|}\sum\limits{{{\left( \hat{\mathbf y}_{p}-{\mathbf y}_{p} \right)}}}^2}$  \\  
-- REL               & $\frac{1}{|\mathbb{P}|}\sum\limits{{{\left| \hat{\mathbf y}_{p}-{\mathbf y}_{p} \right|/{\hat{\mathbf y}_{p}}}}}$  \\ 
-- ${\delta }_{1.25^i}$   & $\frac{|\mathbb{S}|}{|\mathbb{P}|}, \ \mathbb{S}: \max \left( {{\hat{\mathbf y}_p}/{\mathbf y_p},{\mathbf y_p}/{\hat{\mathbf y}_p}} \right)<{1.25}^{i}$  \\ 
\bottomrule
\end{tabular}
}
\vspace{-3pt}
\caption{Metric definition. $\mathbf{y}$: prediction, $\hat{\mathbf{y}}$: ground truth.}\label{tab_metric}
\end{table}

\begin{figure}[t]
\centering
\includegraphics[width=0.9\columnwidth]{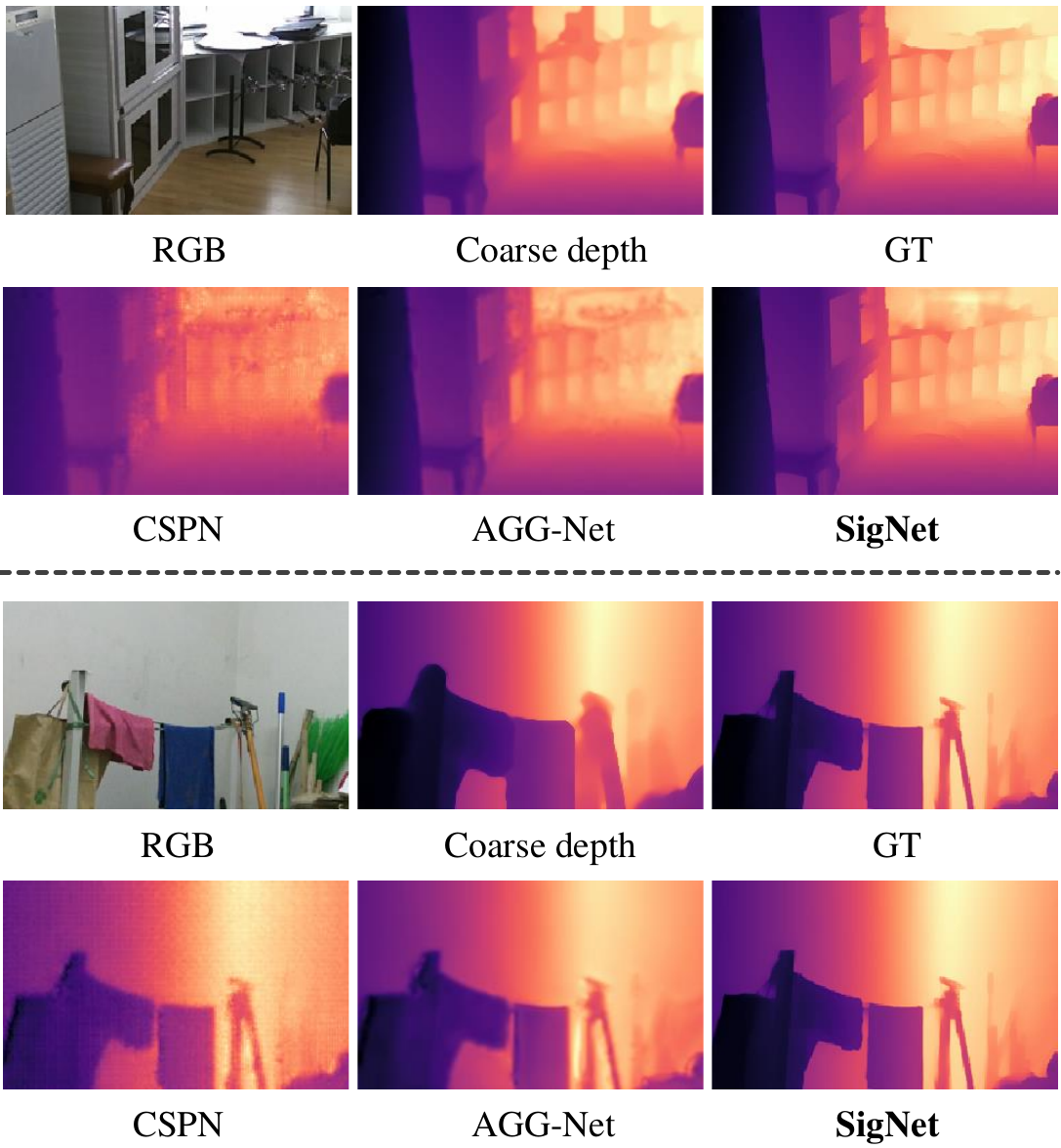}\\
\vspace{-3pt}
\caption{Visual comparisons with SOTAs on DIML.}\label{fig_dimlSupp}
\end{figure}

\section{More Visualizations}
Fig.~\ref{fig_dimlSupp} shows additional visual results on the DIML dataset. As evident, our method reconstructs depth structures with better accuracy and more detailed information. For example, in the first row, the cabinet predicted by our SigNet appears clearer and sharper compared to the other methods, demonstrating the superior performance of our approach in preserving intricate depth details. 

\end{document}